\documentclass{ieeeaccess}
\usepackage{cite}
\usepackage{amsmath,amssymb,amsfonts}
\usepackage{algorithm,algorithmic}
\usepackage{graphicx}
\usepackage{textcomp}
\usepackage{amssymb}
\usepackage{pifont}
\usepackage{array, tabularx, boldline}
\usepackage{hhline}
\usepackage{multirow}
\usepackage{pgfplots,caption}
\usepackage{tikz}
  \NewSpotColorSpace{PANTONE}
  \AddSpotColor{PANTONE} {PANTONE3015C} {PANTONE\SpotSpace 3015\SpotSpace C} {1 0.3 0 0.2}
  \SetPageColorSpace{PANTONE}
\usepackage{ctable}

\usepackage{stackengine}
\usepackage[colorinlistoftodos]{todonotes}

\def\BibTeX{{\rm B\kern-.05em{\sc i\kern-.025em b}\kern-.08em
    T\kern-.1667em\lower.7ex\hbox{E}\kern-.125emX}}
\begin{document}

\title{
Guided Table Structure Recognition through Anchor Optimization}
\author{\uppercase{Khurram Azeem Hashmi}\authorrefmark{1,2,3},     
\uppercase{Didier Stricker\authorrefmark{1,2}, Marcus Liwicki\authorrefmark{4}, Muhammad Noman Afzal\authorrefmark{5} and Muhammad Zeshan Afzal\authorrefmark{1,2,3}}
}

\address[1]{German Research Center for Artificial Intelligence, 67663 Kaiserslautern, Germany}
\address[2]{Department of Computer Science, University of Kaiserslautern, 67663 Kaiserslautern, Germany}
\address[3]{Mindgrage, University of Kaiserslautern, 67663 Kaiserslautern, Germany}
\address[4]{Luleå University of Technology, A3570 Luleå, Sweden}
\address[5]{Bilojix Soft Technologies, Bahawalpur. Pakistan} 

\markboth
{KA Hashmi \headeretal: Table Structure Recognition through Anchor Optimization}
{KA Hashmi \headeretal: Table Structure Recognition through Anchor Optimization}

\corresp{Corresponding author: Khurram Azeem Hashmi (e-mail: Khurram\_Azeem.Hashmi@dfki.de).}

\begin{abstract}

This paper presents the novel approach towards table structure recognition by leveraging the guided anchors. The concept differs from current state-of-the-art approaches for table structure recognition that naively apply object detection methods. In contrast to prior techniques, first, we estimate the viable anchors for table structure recognition. Subsequently, these anchors are exploited to locate the rows and columns in tabular images. Furthermore, the paper introduces a simple and effective method that improves the results by using tabular layouts in realistic scenarios. The proposed method is exhaustively evaluated on the two publicly available datasets of table structure recognition i.e ICDAR-2013 and TabStructDB. We accomplished state-of-the-art results on the ICDAR-2013 dataset with an average F-Measure of 95.05$\%$ (94.6$\%$ for rows and 96.32$\%$ for columns) and surpassed the baseline results on the TabStructDB dataset with an average F-Measure of 94.17$\%$ (94.08$\%$ for rows and 95.06$\%$ for columns).




\end{abstract}



\maketitle

\section{Introduction}
\label{sec:introduction}

In this modern age of digitization, several camera-equipped devices \cite{b60} have been operated daily to upload documents which leads to expanding the need for robust systems that can extract information from raw document images \cite{b61}. In the past, numerous approaches have advertised remarkable results in retrieving information by applying Optical Character Recognition (OCR) methods on documents \cite{b37, b38, b39}. One of the most appropriate ways to represent the information in documents is through tables \cite{b2}. The table contains highly important facts and figures stored in a concise and organized manner \cite{b10}. These tabular structures are extensively used as a medium to convey valuable information in domains like finance, administration, research, and even historical documents \cite{b41}. Hence, automated identification of these tabular structures is a significant problem in the document analysis community \cite{b2,b3,b4}.

The problem of table analysis can be explained by breaking it down into two sub-problems: The first problem is to identify the boundary of the table in a document image whereas the second task is to recognize the structure of the detected table\cite{b10}.

The task of table detection is a complex problem because of the diversity in tabular patterns, for instance, some tables contain ruling lines while others do not have any kind of information. It is highly probable to detect false positives while spotting a table because of having similarities between tables and charts or figures \cite{b48}. These challenges demonstrate that custom heuristics or traditional approaches are not capable of handling the problem of table detection \cite{b41}. The recent development in deep learning based methods has exceptionally improved state-of-the-art table detection methods. Several researchers have exploited deep learning algorithms to detect the tabular area \cite{b23,b28,b40}. Object detection algorithms have been proven to surpass the rest of the techniques and achieved almost perfect results \cite{b41,b56}.

\begin{figure}[ht]
    \includegraphics[width =\columnwidth]{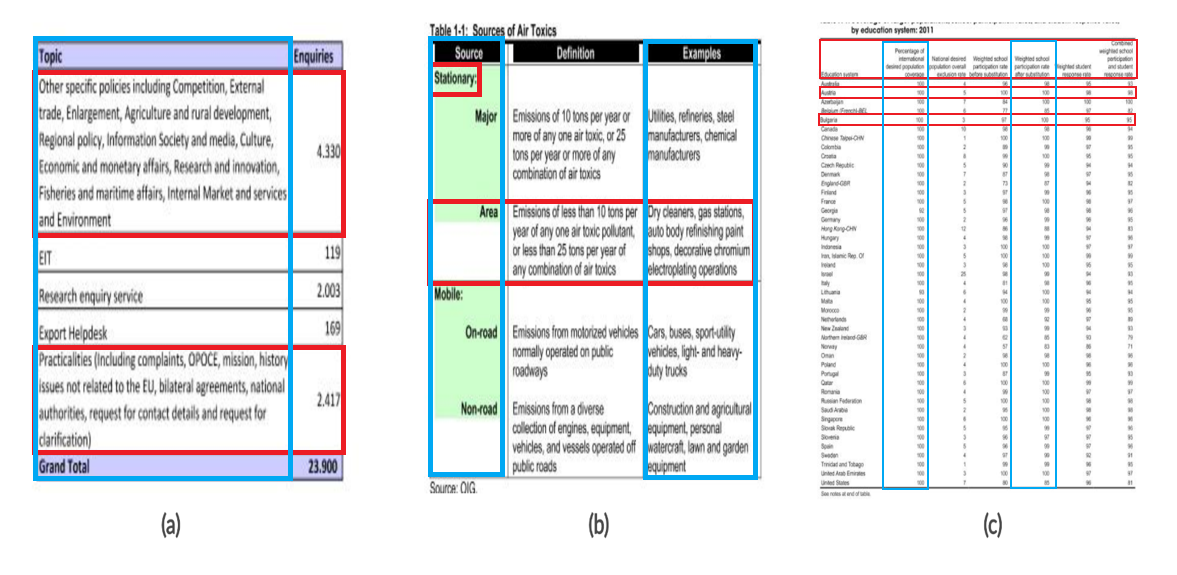}
    \caption{Table Structure Recognition problem definition and challenges. Red color defines the bounding box for rows while blue color denotes columns. In the figure, part(a) and part(b) represents tabular images having rows spanning multiple lines whereas, in part(c), rows are restricted to a single line. Columns can be as wide as illustrated in part(a) and part(b) but also as narrow as shown in part(c). For the sake of clarity, only a few rows and columns are highlighted.}
    \label{fig:table_str_confusion}
\end{figure}

The task of table structure recognition is the detection of various cells present in the table \cite{b18}. This problem can be further dissolved into detecting rows and columns in a table that can be later combined to produce the respective cells \cite{b27}. The pre-condition for the task of table structure recognition is the accurately detected tabular regions \cite{b33,b32}. Figure \ref{fig:table_str_confusion} illustrates how the problem of table structure recognition is defined in our approach. Additionally, the figure depicts the challenges that exist due to the diversities in structures of rows (columns) in tabular images. Only few rows (columns) are marked for the sake of clarity.

There are several approaches that have tackled the problem of table structure recognition by leveraging additional metadata extracted from the PDF files \cite{b14,b19}. However, extracting tabular structures directly from images is perplexing as compared to operating over digital-born PDFs \cite{b27}. Although few considerable efforts have also been made to recognize the tabular structures straight from images 
\cite{b28,b32}, accurate structural recognition is far from achievable \cite{b33}.

This paper extends the idea of treating the problem of table structure recognition as an object detection problem \cite{b33}. In object detection problems, the elementary task is to find the object in a natural scene image. In our case, we operate a document as a natural image while the rows and columns in the table are our targeted objects. While the system DeepTabStr \cite{b33} relied on memory-intensive deformable convolutions \cite{b42}, our approach consists of intuitive utilization of Mask R-CNN \cite{b36} with optimized anchors along with a simple and effective post-processing method. 

In particular, the contributions of this paper are summarized as follows:
\begin{itemize}
  \item We have treated the problem of \textbf{table structure extraction as an object detection problem} by employing the well-known Mask R-CNN model.
  \item We have implemented a novel \textbf{anchor optimization technique} in a region-based convolutional neural network that produces faster network convergence. 
  \item We have introduced a simple and effective \textbf{post-processing method to remove the extra white spaces} from the predicted rows. This method can be exploited to recognize tabular structures in realistic scenarios.
  \item After \textbf{extensive cross-dataset evaluations}, our proposed approach has beaten the \textbf{state-of-the-art results on the ICDAR-13 dataset} \cite{b18} by using same evaluation metrics proposed by Schreiber \textit{et al.} \cite{b27}. Furthermore, we have also \textbf{surpassed the baseline results on the TabStructDB} dataset \cite{b33}.
\end{itemize}

The rest of the paper is organized as follows: In the beginning, we discuss some of the previous work closely related to our approach in Section \ref{sec:related_work}; In Section \ref{sec:methodology} we explain our proposed approach and discuss the ideas used in the experiments; Section \ref{sec:datasets} provides a brief overview about the datasets which are exploited in the proposed approach; Along with a brief detail over evaluation metrics, we present our results in Section \ref{sec:evaluation}; Finally, Section \ref{sec:conclusion} concludes the paper.

\section{RELATED WORK}
\label{sec:related_work}

In this section, we highlight the most relevant related work in field of table structure analysis. The contributions can be divided into pre and post-deep learning era as described in the following sections. For an exhaustive state-of-the-art overview in the closely related research area of table understanding, refer to \cite{b1,b2,b3,b4,b5,b6,b7,b8,b9,b10}.


\begin{figure*}[ht]
    \includegraphics[width =\linewidth]{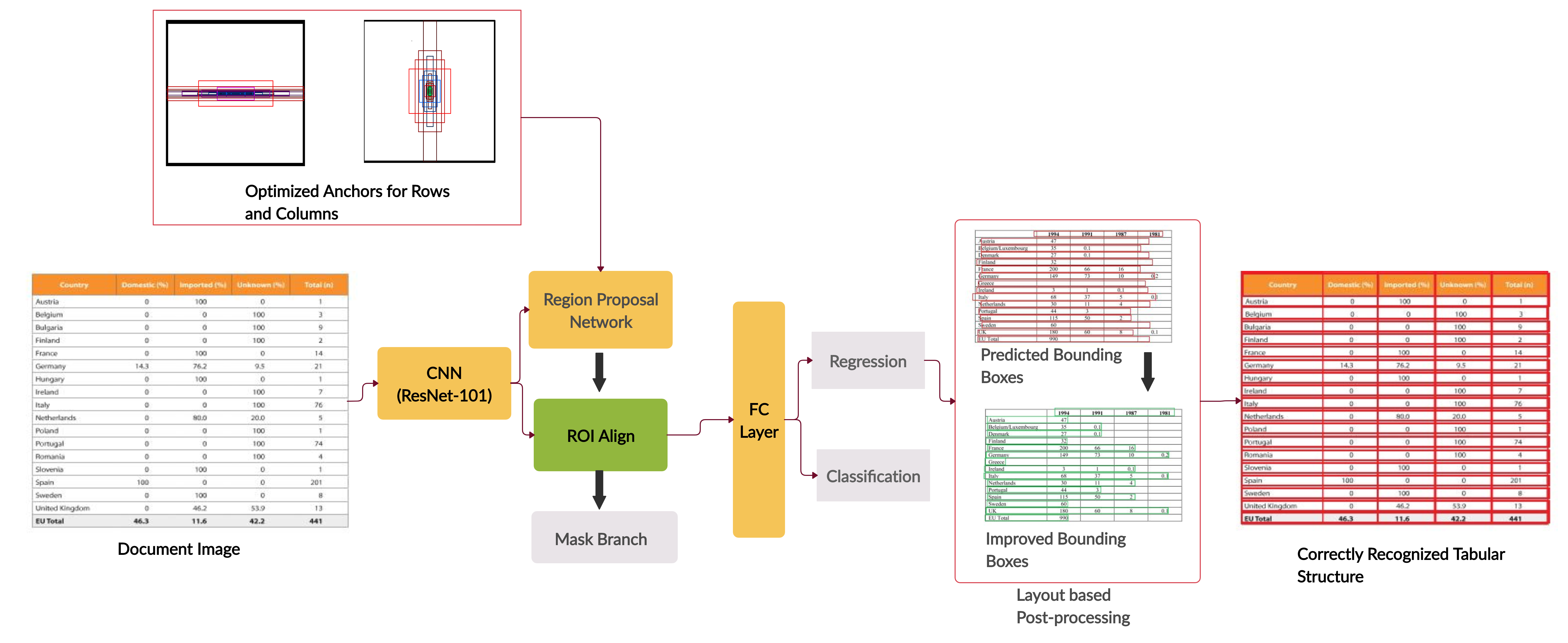}
    \caption{The proposed pipeline for Table Structure Recognition. Optimized anchors are given to the region proposal network of Mask R-CNN. After regressing coordinates by the network, the predicted bounding boxes for row detection are further enhanced by employing the post-processing technique.}
    \label{fig:architecture_diagram}
\end{figure*}

\subsection{Traditional Approaches}
Kieninger \textit{et al.} \cite{b11,b12,b13}, who are the pioneers for working in table structure extraction, tackled the problem by leveraging the traditional approaches. Their proposed system T-Recs gathered the words into columns by calculating their horizontal ruling lines. Subsequently, horizontal lines were split into respective cells based on column margins.

Wang \textit{et al.} \cite{b15} proposed a system which can generate a large number of table ground truths which are beneficial for table recognition systems. The author used a novel table analysis algorithm along with an X-Y cut algorithm to extract table structure by detecting the respective cells. Later, another data-driven system proposed by Wang \textit{et al.} \cite{b16} which was based on joint probability distributions and deals with both detection and structure decomposition of tables. Their algorithm was analogous to a well famous X-Y cut algorithm \cite{b17}.

The problem of table structure extraction caught attention when a table structure recognition competition is organized at ICDAR in 2013 \cite{b18}. While the first part of the competition was to detect the boundary of the table, the second part of this competition was to recognize the tabular structure by reconstructing the cellular structure of a table. The cell-level metrics were used to evaluate the performance of the systems. It is important to note that apart from one candidate, all of the participants in the competition vastly used the PDF metadata. However poor results achieved by the pure image-based system depict that cell-level metrics are not suitable for the evaluation of image-based table analysis systems.

Another approach that leverages PDF-metadata to detect the structure of tables is published by Klampfl \textit{et al.} \cite{b19}. The system employed the blend of unsupervised learning techniques and hand-crafted heuristics to perform table structure recognition. Kasar \textit{et al.} \cite{b20} came up with a query-based system to extract structure of the tables. The system converts the input query taken from the user into a relational graph which is then compared using a graph matching algorithm to fetch the required information.

Shigarov \textit{et al.} \cite{b21} performed an exhaustive evaluation on various algorithms with different thresholds and custom heuristics to tackle the problem of table structure recognition. Their approach was heavily dependent on the PDF meta-data as well. Another approach relying heavily on PDF-metadata is proposed by Rastan \textit{et al.} \cite{b22}. Along with recognizing the structure of tables, their system TEXUS can also extract the content from tabular structures.

All these techniques are heavily dependent on the meta-data available in digital-born PDFs. Since our approach works on the scanned document images, these techniques are not directly comparable with our approach.

\subsection{Deep Learning based Approaches}

\subsubsection{Graph Neural Networks}
Recently, Chi \textit{et al.} \cite{b23} has exploited graph neural networks \cite{b28} to perform the task of table structure recognition on PDF documents. Another approach powered by graph neural networks is published by Qasim \textit{et al.} \cite{b24}. Their model combines the capabilities of convolutional neural networks and graph neural networks to extract tabular structures. Xue \textit{et al.} introduced a bottom-up approach by reconstructing the table structure using a cell relationship network. The system ReS${^2}$TIM \cite{b25} employed a distance-based weight technique to retrieve a syntactic table structure.

\subsubsection{Recurrent Neural Networks}

Recurrent neural networks \cite{b29} have also been employed to handle the problem of table structure extraction \cite{b30,b31}. However, most of the prior approaches have utilized PDF meta-data. Since we deal with natural document images, they are not directly comparable to our approach.

\subsubsection{Convolutional Neural Networks}

To the best of our knowledge, Schreiber \textit{et al.} \cite{b27} published the first natural image-based deep learning system which explored the problem of table structure analysis. The system leverages the Fully Convolutional Network (FCN) \cite{b62} to segment the table into rows and columns. Later in 2019, TableNet has been proposed by Shubham \textit{et al.} \cite{b28}. The authors tackled the problem of table structure extraction through a semantic segmentation technique. Another approach powered by semantic segmentation to extract tabular structures from document images is published by Siddiqui \textit{et al.} \cite{b32}.

\begin{figure*}[ht]
    \includegraphics[width =\linewidth]{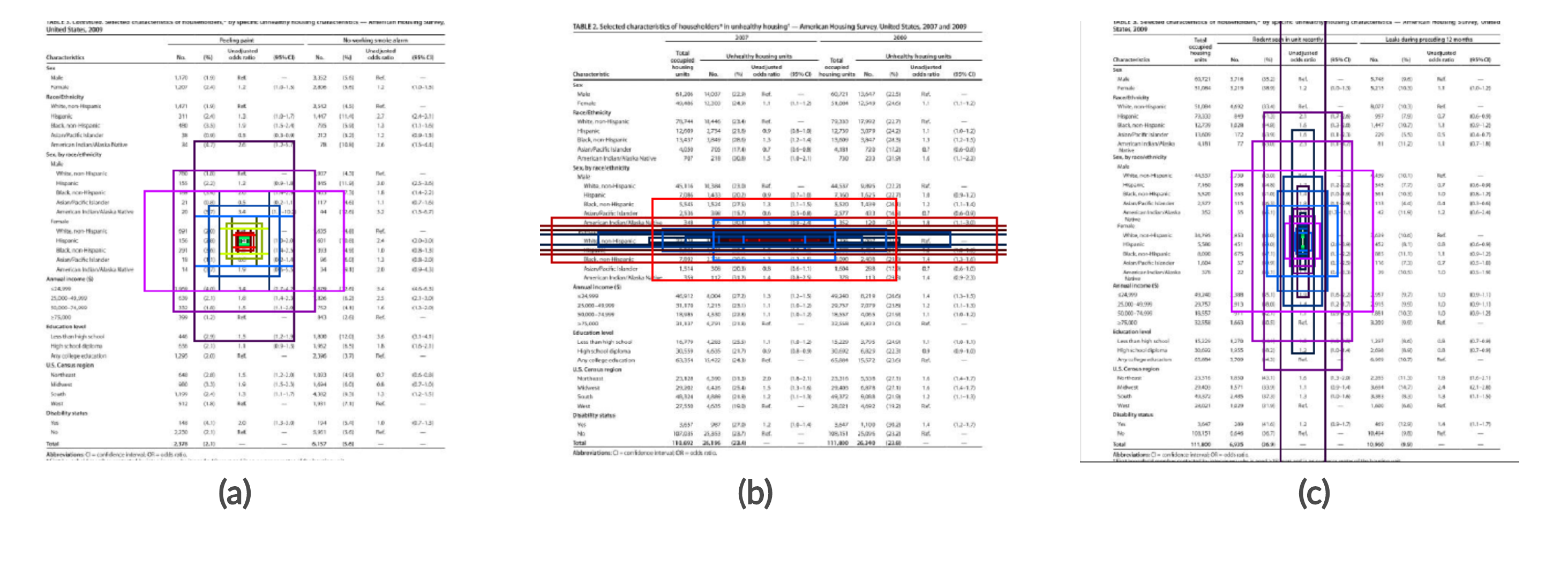}
    \caption{Visualisation of anchors traditionally used for object detection techniques against optimized anchors used in our approach. (a)
    Anchors traditionally used for object detection. (b) Optimized anchors for row detection. (c) Optimized anchors for column detection. Traditional anchors are transformed into optimized anchors using K-Means Clustering technique. }
    \label{fig:anchors_comprison}
\end{figure*}

All of these approaches have either used semantic segmentation or FCN to solve the problem of document images. Contrarily, we have chosen to handle the task of table structure recognition as an object detection problem. Although Siddiqui \textit{et al.} \cite{b33} has treated the table structure analysis as an object detection problem, there are various considerable differences between the two methods. The system DeepTabStr \cite{b33} has adopted Faster R-CNN \cite{b34} with deformable convolutions \cite{b42} while our proposed approach works with Mask R-CNN \cite{b36} exploiting optimized anchors to directly detect boundaries of respective rows and columns in a tabular image.

\section{METHOD}
\label{sec:methodology}
We have devised the problem of table structure recognition as an object detection problem. Object detection is a famous problem in computer vision that studies how an object can be recognized from a natural scene image. Recent progress in deep learning has remarkably enhanced the state-of-the-art object detection systems \cite{b34, b36}. To achieve the ultimate goal of table structure recognition, we have decomposed our problem into two sub-problems where the first one is about detecting rows in tables while the second sub-problem deals in detecting columns.

\subsection{Model}

Our proposed approach could be implemented in two ways: 
\begin{enumerate}
    \item Separate model for both rows and columns.
    \item Single combined model to handle both the problems. 
\end{enumerate}

Considering the diversity in the structures of rows and columns, it has been settled that the separate model performs better \cite{b33}. Hence, we have decided to go for two segregated models to solve the problem of table structure recognition.

We have adopted Mask R-CNN \cite{b36} as our model to identify the rows and columns in a table. Mask R-CNN is one of the accurate object detection algorithms and the latest member to the group of Region-based Convolutional Neural Networks (RCNN) \cite{b44}. Mask R-CNN is a two-phase model and has shown compelling performance on the PASCAL VOC \cite{b45} and COCO \cite{b46} datasets. Mask R-CNN has been exploited before to identify various graphical objects in document images \cite{b48}. 

To execute the training process of the deep neural network, it requires an extensive amount of data which we lack specifically in the domain of table structure extraction. To tackle this problem, we have leveraged the capabilities of transfer learning in our approach. The backbone of our Mask R-CNN is a  pre-trained model on ImageNet \cite{b47} dataset. 

Figure \ref{fig:architecture_diagram} illustrates the complete pipeline of our proposed approach. Analogous to Faster R-CNN \cite{b34}, Mask R-CNN \cite{b36} follows the two-phase procedure with one addition. The first phase consists of Region Proposal Network (RPN) which proposes regions of interest in a document image whereas the second phase deals in the classification of labels and regression of bounding boxes including the binary masks of each region of interest. Now, we will discuss in detail about the different components of our proposed pipeline presented in Figure \ref{fig:architecture_diagram}.

In the first stage, the combination of ResNet-101 \cite{b49} and Feature Pyramid Network (FPN) \cite{b50} which is acting as a backbone in our case, extracts the features from the document image. These features are further propagated to Region Proposal Network (RPN). RPN is a lightweight neural network that scans some regions in an image and tries to filter out the ones which are more likely to contain objects. These input regions for RPN are known as \textit{anchors}. Anchors are defined as a set of rectangular regions with a predefined set of scales and aspect ratios \cite{b59}. The RPN generates two kinds of outputs for each anchor: 
\begin{enumerate}
    \item The class of an anchor which states whether an anchor is an object or background.
    \item Bounding box refinement which is the change in the position of the bounding box to precisely fit the object in the proposed region of interest.
\end{enumerate}

\subsection{Anchor Optimization}
The concept of anchors was introduced in the Faster R-CNN by Ren \textit{et al.} \cite{b34} which is transported into Mask R-CNN \cite{b36} as well. Contrary to the hand-crafted approach of selecting anchors in Mask R-CNN, we have applied the K-means clustering technique to retrieve fine anchors as explained in the approach proposed by Redmon \textit{et al.} \cite{b51}. The anchors traditionally used in object detection consist of various width to height ratios to deal with objects having diverse shapes \cite{b63}. However, in the case of detecting rows, we are aware that the width of the anchor will always be equal or greater than the height of an anchor while it is the other way around for columns. Hence, anchors having customized sizes and aspect ratios will lead to better performance as compared to anchors commonly used for object detection techniques. It is important to mention that the euclidean distance  was not used as a distance metric in our K-means clustering technique but the following distance metric \cite{b56} is used :
\begin{equation}
\label{eq:dist}
D(box,centroid) = 1 - IoU(box,centroid)
\end{equation}

where the \textit{box} represents bounding box as a data sample that needs to be clustered and \textit{centroid} is the center of a cluster which will be retrieved at the end of clustering. IoU (Intersection over Union) is an evaluation metrics which is explained in Section \ref{sec:evaluation}. The purpose of choosing this metric over euclidean distance is that the bigger boxes will lead to more errors as compared to smaller ones which is not the main concern in our scenario \cite{b56}. The traditional anchors are given as input to K-means clustering technique along with the training datasets of ICDAR 2013 \cite{b18} and TabStructDB  \cite{b33} in order to retrieve optimized anchors for each dataset. 

Figure \ref{fig:anchors_comprison} illustrates the comparison between original anchors used for object detection and the optimized anchors for the row (column) detection. The anchor ratios (0.5, 1 and 2) are used in Figure \ref{fig:anchors_comprison}(a) while for Figure \ref{fig:anchors_comprison}(b) and Figure \ref{fig:anchors_comprison}(c), we have used four different anchor ratios (50, 25, 10 and 3) and (0.1, 0.3, 0.5 and 1) respectively. It can be perceived that optimized anchors \ref{fig:anchors_comprison}(b) and \ref{fig:anchors_comprison}(c)  are well suited to execute the task of table structure recognition as compared to the anchors traditionally used for object detection \ref{fig:anchors_comprison}(a) .

\begin{figure}[ht]
    \includegraphics[width =\columnwidth]{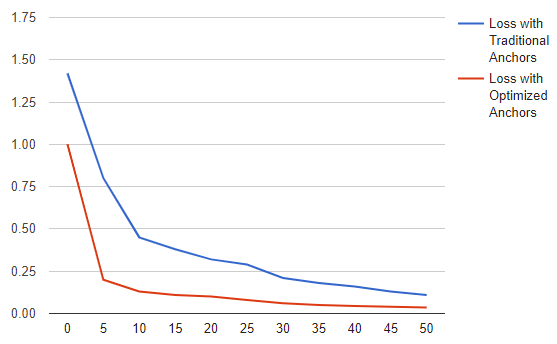}
    \caption{Network training loss comparison between Optimized anchors and traditional anchors for row detection. Blue line chart represents network training loss for row detection using traditional anchors. Red line chart represents network training loss for row detection using optimized anchors. The Y axis of the graph shows the loss values whereas the X axis displays the number of epochs. It can be seen that the network with optimized anchors achieves the loss value of less than 0.1 right after the 30 epochs whereas the network with traditional anchors is unable to achieve the same loss value even after 50 epochs.}
    \label{fig:anchor_network}
\end{figure}

\begin{figure}[ht]
    \includegraphics[width =\columnwidth]{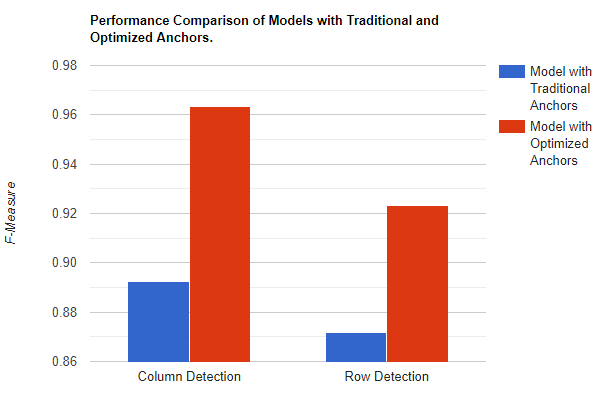}
    \caption{Performance comparison between the models trained with traditional anchors and optimized anchors. We have experienced a noticeable increase in the F-measure score for both rows and columns detection after employing optimized anchors.}
    \label{fig:model_perf_anchor_op}
\end{figure}

It is important to mention that RPN scans these optimized anchors on the feature maps instead of an actual document image. This enables the RPN to reuse extracted features efficiently. The proposed anchor optimization technique not only improves the performance of the model but also facilitates the network to converge faster, making our approach even more efficient. Figure \ref{fig:anchor_network} illustrates how optimized anchors help the network to achieve better results in less time. Along with faster network convergence, the performance of our model is significantly improved after exploiting optimized anchors. The performance comparisons between the models for row and column detection employed with traditional and optimized anchors are explained in Figure \ref{fig:model_perf_anchor_op} and their F-Measure scores are summarized in Table \ref{tab:anchor_result}.


\begin{table}
    \centering
    \normalsize
    \setlength\tabcolsep{5pt} 
    \setlength\extrarowheight{5pt}
    \begin{tabular}{|c|c|c|} 
    \hline  
    \normalsize {Model} &
    \normalsize  {Row Detection}&
    \normalsize  {Column Detection}\\
    \hline  
    \small {Traditional Anchors} &
    \small  {0.8710}&
    \small  {0.8920}\\
    \hline  
    \small \textbf{Optimized Anchors} &
    \small  \textbf{0.9206}&
    \small  \textbf{0.9632}\\
    
    \hline
\end{tabular}
    \caption{Comparison of F-Measure scores for rows and column detection between traditionally used anchors and our optimized anchors.}
    \label{tab:anchor_result}
\end{table}


\subsection{Layout Based Post Processing} 

Once the rows and columns are detected by the Mask R-CNN, we noticed that while the network managed to detect the columns properly, it was unable to recognize the precise boundaries of rows. In the case of row detection, we observed that the height of predicted bounding boxes is identical to ground truth, however, the network struggled to predict the accurate width of a bounding box. This either generates extra white spaces in the bounding box or drops some valuable information from the rows. To tackle the problem, we came up with a simple and effective post-processing algorithm that can resize the width of a bounding box on the basis of few constraints.

We are aware of the fact that for most of the documents, the information is written in black color. Our proposed method improves precision and recall in two ways:

\begin{enumerate}
    \item Incorporating the important information which was overlooked by the network by increasing the width of a bounding box close to the last set of black pixels.
    \item Removing extra white spaces by decreasing the width of the bounding box to the nearest set of black pixels.
\end{enumerate}

\begin{figure*}[ht]
    \includegraphics[width =\linewidth]{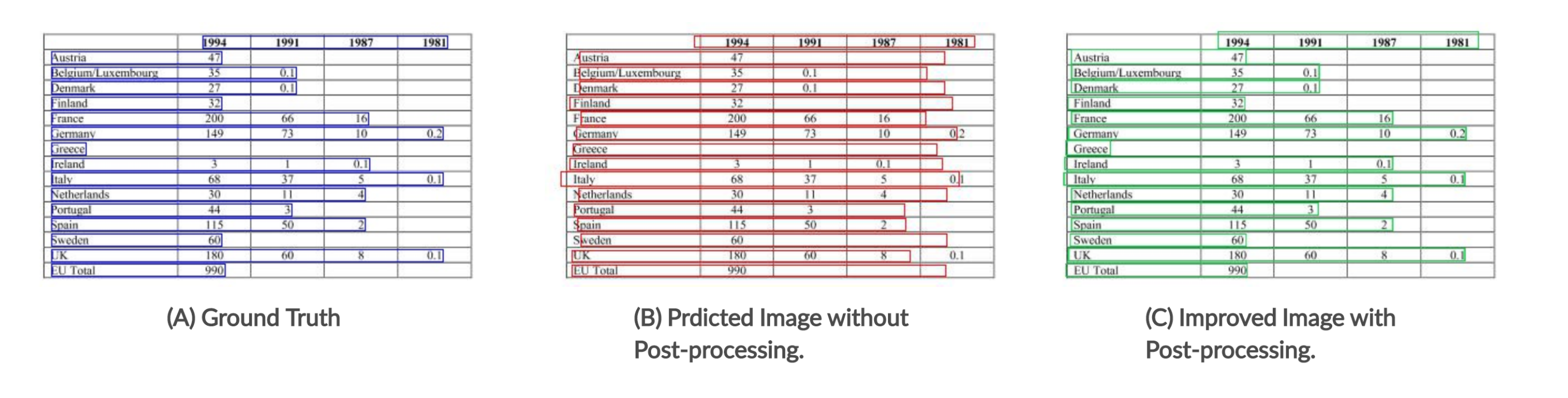}
    \caption{Explaining through example about how the IoU for row detection in document images can be further improved with simple post-processing. Detected rows in part (B) are either stretched or reduced to produce accurate boundaries as illustrated in part (C).}
    \label{fig:post_processing_results}
\end{figure*}

The pseudo-code for the proposed method is explained in Algorithm \ref{algo}. It is important to mention that method does not work in a few of the cases where the text is not displayed in black pixels. However, the proposed method has shown significant improvements in the IoU of bounding boxes in case of row detection which is summarized in Table \ref{tab:post_process_results}. Figure \ref{fig:post_processing_results} depicts how the performance of row detection can be improved from simple post-processing.

\renewcommand{\algorithmicrequire}{\textbf{Input: }\textit{I}: 2d array of predicted bounding box}
\renewcommand{\algorithmicensure}{\textbf{Output: }\textit{R}: Improved bounding box}

\begin{algorithm}

  \caption{Resize the width of bounding box by identifying black pixels}
  \begin{algorithmic}[1]
    \REQUIRE 
    \ENSURE
     \STATE $blackPT \gets  Black Pixel Threshold $
     \STATE $Area \gets  Image Specfic Area $
      \STATE $R \gets I$
      
      \FOR{$xValue$ of $R$ to end of image} \COMMENT{checking forward for both xmin and xmax}
        \IF{$blackPixel found $}
            \STATE Compute blackPixel count in that $Area$
            \IF{$blackPixelCount >= blackPT $}
                \STATE $xmax of R \gets xValue$
            \ENDIF
        \ENDIF
      \ENDFOR
      \FOR{$xValue$ of $R$ to beginning of image} \COMMENT{checking backward for both xmin and xmax} 
        \IF{$blackPixel found $}
            \STATE Compute blackPixel count in that $Area$
            \IF{$blackPixelCount >= blackPT $}
                \STATE $xmin of R \gets xValue$
            \ENDIF
        \ENDIF
      \ENDFOR
    \STATE return $R$
  \end{algorithmic}
\label{algo}
\end{algorithm}

\subsection{Experiment Details}

We have worked with the combination of ResNet-101 \cite{b49} and FPN\cite{b50} model as a backbone for both the row and columns detection. Apart from the aspect ratios anchors, the rest of the hyperparameters were identical for both of the models. For row detection, we have used four different anchor ratios (50, 25, 10, and 3) whereas, for columns, we have picked four different anchors ratios with (0.1, 0.3, 0.5, and 1). However, we have used five different anchor scales (16, 32, 64, 128, 256) for both of the networks. Both of the models were optimized for 50 epochs where each epoch consists of 100-time steps. The maximum image size was limited to $1024 \times 800$ and the images exceeding this size were resized to the maximum dimension. We used a batch size of 2 on a single NVIDIA $1080$ Ti GPU. Our model works on stochastic gradient descent having a momentum value of 0.9 and a learning rate of 0.0001. Gradients are clipped to 5.0 and weights are decayed by 0.0001 at each epoch. In order to prevent the problem of overfitting, we have applied augmentation techniques like random rotations, Gaussian blurring, and random horizontal and vertical flips on the training dataset. We have implemented this work in Keras \cite{b53} with Tensorflow \cite{b54} as a backend.

\begin{table}
    \centering
    \normalsize
    \setlength\tabcolsep{5pt} 
    \setlength\extrarowheight{5pt}
\begin{tabular}{ |c|c|c|c| } 
    \hline
    \normalsize {Approach} &
    \normalsize  {Precision}&
    \normalsize  {Recall}&
    \normalsize  {F-Measure} \\
    \hline
    \small {Before Post-processing} &
    \small  {0.9106}&
    \small  {0.9326}&
    \small {0.9206} \\
    \hline
    \small {After Post-processing} &
    \small  \textbf{{0.9468}}&
    \small  \textbf{{0.9452}}&
    \small \textbf{{0.9460}} \\
    \hline
\end{tabular}
    \caption{Performance comparison for row detection before and after applying post-processing technique. After applying our post-processing method, we have seen a significant increase in an F-Measure score in case of row detection in document images.}
    \label{tab:post_process_results}
\end{table}


\newcolumntype{d}{X}
\newcolumntype{u}{>{\arraybackslash\hsize=.5\hsize}X}
\newcolumntype{v}{>{\hsize=.4\hsize}X}
\newcolumntype{t}{>{\hsize=.65\hsize}X}
\newcolumntype{k}{>{\hsize=.55\hsize}X}{c{2cm}}

\begin{table*}
    \centering
    \setlength\tabcolsep{5pt} 
    \setlength\extrarowheight{5pt}
    
    \begin{tabularx}{\linewidth}{|u|u|t|v|v|k|v|v|k|k|}
        \specialrule{.2em}{.1em}{.1em} 
        \captionsetup{font=scriptsize}
        \multirow{2}{2cm}{\textbf{Training Dataset}}&
        \multirow{2}{2cm}{\textbf{Testing \newline Dataset}}&
        \multirow{2}{2cm}{\textbf{Model}}&
        \multicolumn{3}{c|}{\textbf{Row}}&
        \multicolumn{3}{c|}{\textbf{Column}}&
        \multirow{2}{2cm}{\textbf{Average F-Measure}}\\
        \cline{4-9}
        & & & \footnotesize \textbf{Precision}&
        \footnotesize \textbf{Recall}&
        \footnotesize \textbf{F-Measure}&
        \footnotesize \textbf{Precision}&
        \footnotesize \textbf{Recall}&
        \footnotesize \textbf{F-Measure}&
        \small \textbf{ }\\
        \hline
        \footnotesize \multirow{3}{2cm} {\newline \newline \newline\newline \newline \newline \newline \newline \newline \newline \newline ICDAR-13 \newline (Training set)}&
        \footnotesize \multirow{4}{2cm} {\newline ICDAR-13 \newline (Test set)}&
        \footnotesize  Faster R-CNN&
        
        \footnotesize \centering {0.8974}&
        \footnotesize \centering {0.9154}&
        \footnotesize \centering {0.9063}&
        \footnotesize \centering {0.9456}&
        \footnotesize \centering {0.9488}&
        \footnotesize \centering {0.9510}&
        \footnotesize 
        {0.9286} \\
         \cline{3-10}
        & &\footnotesize  Deformable Faster R-CNN &
        
        \footnotesize \centering {0.9071}&
        \footnotesize \centering {0.9221}&
        \footnotesize \centering {0.9145}&
        \footnotesize \centering {0.9511}&
        \footnotesize \centering {0.9588}&
        \footnotesize \centering {0.9549}&
        \footnotesize {0.9347} \\
         \cline{3-10}
         & &\footnotesize  Deformable Faster R-CNN$\dagger$&
        \footnotesize \centering {0.8817}&
        \footnotesize \centering {0.4097}&
        \footnotesize \centering {0.4531}&
        \footnotesize \centering {0.9520}&
        \footnotesize \centering {0.9497}&
        \footnotesize \centering {0.9497}&
        \footnotesize {0.7014} \\
         \cline{3-10}
        & &\footnotesize  Mask R-CNN&
        \footnotesize \centering \textbf{0.9106}&
        \footnotesize \centering \textbf{0.9326}&
        \footnotesize \centering \textbf{0.9206}&
        \footnotesize \centering \textbf{0.9605}&
        \footnotesize \centering \textbf{0.9659}&
        \footnotesize \centering \textbf{0.9632}&
        \footnotesize \textbf{0.9419} \\
        \cline{2-10}
        & \footnotesize \multirow{4}{2cm}{TabStructDB \newline (Complete)}&
        \footnotesize  Faster R-CNN&
        \footnotesize \centering {0.6968}&
        \footnotesize \centering {0.6632}&
        \footnotesize \centering {0.6796}&
        \footnotesize \centering {0.6845}&
        \footnotesize \centering {0.6973}&
        \footnotesize \centering {0.6908}&
        \footnotesize 
        {0.6852} \\
         \cline{3-10}
        & &\footnotesize  Deformable Faster R-CNN &
        \footnotesize \centering {0.7034}&
        \footnotesize \centering \textbf{0.6972}&
        \footnotesize \centering {0.7003}&
        \footnotesize \centering \textbf{0.7883}&
        \footnotesize \centering \textbf{0.7561}&
        \footnotesize \centering \textbf{0.7719}&
        \footnotesize \textbf{0.7361} \\
         \cline{3-10}
         & &\footnotesize  Deformable Faster R-CNN$\dagger$&
        \footnotesize \centering {0.5545}&
        \footnotesize \centering {0.2785}&
        \footnotesize \centering {0.4531}&
        \footnotesize \centering {0.7681}&
        \footnotesize \centering {0.7489}&
        \footnotesize \centering {0.7533}&
        \footnotesize {0.6032} \\
         \cline{3-10}
        & &\footnotesize  Mask R-CNN&
        \footnotesize \centering \textbf{0.7189}&
        \footnotesize \centering {0.6850}&
        \footnotesize \centering \textbf{0.7034}&
        \footnotesize \centering {0.7037}&
        \footnotesize \centering {0.7157}&
        \footnotesize \centering {0.7097}&
        \footnotesize 
        {0.7011} \\
        \cline{2-10}
        
        & \footnotesize \multirow{4}{2cm}{TabStructDB (Test Set)}&
        \footnotesize  Faster R-CNN&
        \footnotesize \centering {0.6788}&
        \footnotesize \centering {0.6634}&
        \footnotesize \centering {0.6710}&
        \footnotesize \centering {0.7041}&
        \footnotesize \centering {0.7255}&
        \footnotesize \centering {0.7146}&
        \footnotesize 
        {0.6928} \\
         \cline{3-10}
        & &\footnotesize  Deformable Faster R-CNN &
        
        \footnotesize \centering {0.6925}&
        \footnotesize \centering {0.6812}&
        \footnotesize \centering {0.6868}&
        \footnotesize \centering {0.7152}&
        \footnotesize \centering {0.7377}&
        \footnotesize \centering {0.7263}&
        \footnotesize {0.7066} \\
         \cline{3-10}
         & &\footnotesize  Deformable Faster R-CNN$\dagger$&
        \footnotesize \centering {0.5492}&
        \footnotesize \centering {0.2622}&
        \footnotesize \centering {0.3009}&
        \footnotesize \centering \textbf{0.7687}&
        \footnotesize \centering {0.7462}&
        \footnotesize \centering \textbf{0.7501}&
        \footnotesize {0.5255} \\
         \cline{3-10}
        & &\footnotesize  Mask R-CNN&
        \footnotesize \centering \textbf{0.7142}&
        \footnotesize \centering \textbf{0.6937}&
        \footnotesize \centering \textbf{0.7039}&
        \footnotesize \centering {0.7376}&
        \footnotesize \centering \textbf{0.7525}&
        \footnotesize \centering {0.7484}&
        \footnotesize 
        \textbf{0.7237} \\
        \cline{2-10}
        \hline
        
        \footnotesize \multirow{3}{2cm} {\newline \newline \newline\newline \newline \newline \newline \newline \newline \newline \newline TabStructDB \newline (Training set)}&
        \footnotesize \multirow{4}{2cm} {\newline ICDAR-13 \newline (Complete)}&
        \footnotesize  Faster R-CNN&
        
        \footnotesize \centering {0.7577}&
        \footnotesize \centering {0.7322}&
        \footnotesize \centering {0.7447}&
        \footnotesize \centering {0.6954}&
        \footnotesize \centering {0.7125}&
        \footnotesize \centering {0.7038}&
        \footnotesize 
        {0.7242} \\
         \cline{3-10}
        & &\footnotesize  Deformable Faster R-CNN &
        
        \footnotesize \centering {0.7821}&
        \footnotesize \centering {0.7514}&
        \footnotesize \centering {0.7664}&
        \footnotesize \centering {0.7023}&
        \footnotesize \centering {0.7344}&
        \footnotesize \centering {0.7180}&
        \footnotesize {0.7422} \\
         \cline{3-10}
         & &\footnotesize  Deformable Faster R-CNN$\dagger$&
        \footnotesize \centering {0.6048}&
        \footnotesize \centering {0.5507}&
        \footnotesize \centering {0.5660}&
        \footnotesize \centering \textbf{0.7308}&
        \footnotesize \centering \textbf{0.7518}&
        \footnotesize \centering \textbf{0.7422}&
        \footnotesize {0.6541} \\
         \cline{3-10}
        & &\footnotesize  Mask R-CNN&
        \footnotesize \centering \textbf{0.8263}&
        \footnotesize \centering \textbf{0.7729}&
        \footnotesize \centering \textbf{0.7987}&
        \footnotesize \centering {0.7143}&
        \footnotesize \centering {0.7226}&
        \footnotesize \centering {0.7184}&
        \footnotesize \textbf{0.7677} \\
        \cline{2-10}
        & \footnotesize \multirow{4}{2cm}{ICDAR-13 \newline (Test Set)}&
        \footnotesize  Faster R-CNN&
        \footnotesize \centering {0.7321}&
        \footnotesize \centering {0.7144}&
        \footnotesize \centering {0.7231}&
        \footnotesize \centering {0.6543}&
        \footnotesize \centering {0.6411}&
        \footnotesize \centering {0.6476}&
        \footnotesize 
        {0.6853} \\
         \cline{3-10}
        & &\footnotesize  Deformable Faster R-CNN &
        
        \footnotesize \centering {0.7932}&
        \footnotesize \centering {0.6350}&
        \footnotesize \centering {0.7053}&
        \footnotesize \centering {0.6621}&
        \footnotesize \centering {0.6721}&
        \footnotesize \centering {0.6671}&
        \footnotesize {0.6862} \\
         \cline{3-10}
         & &\footnotesize  Deformable Faster R-CNN$\dagger$&
        \footnotesize \centering {0.5279}&
        \footnotesize \centering {0.4625}&
        \footnotesize \centering {0.4818}&
        \footnotesize \centering \textbf{0.6701}&
        \footnotesize \centering \textbf{0.6768}&
        \footnotesize \centering \textbf{0.6705}&
        \footnotesize {0.5761} \\
         \cline{3-10}
        & &\footnotesize  Mask R-CNN&
        \footnotesize \centering \textbf{0.8296}&
        \footnotesize \centering \textbf{0.7586}&
        \footnotesize \centering \textbf{0.7925}&
        \footnotesize \centering {0.6453}&
        \footnotesize \centering {0.6307}&
        \footnotesize \centering {0.6379}&
        \footnotesize 
        \textbf{0.7354} \\
        \cline{2-10}
        
        & \footnotesize \multirow{4}{2cm}{TabStructDB (Test Set)}&
        \footnotesize  Faster R-CNN&
        \footnotesize \centering {0.9074}&
        \footnotesize \centering {0.9254}&
        \footnotesize \centering {0.9163}&
        \footnotesize \centering {0.9556}&
        \footnotesize \centering {0.9588}&
        \footnotesize \centering {0.9572}&
        \footnotesize 
        {0.9367} \\
         \cline{3-10}
        & &\footnotesize  Deformable Faster R-CNN &
        
        \footnotesize \centering {0.9111}&
        \footnotesize \centering {0.9285}&
        \footnotesize \centering {0.9197}&
        \footnotesize \centering \textbf{0.9605}&
        \footnotesize \centering {0.9659}&
        \footnotesize \centering \textbf{0.9632}&
        \footnotesize {0.9414} \\
         \cline{3-10}
         & &\footnotesize  Deformable Faster R-CNN$\dagger$&
        \footnotesize \centering {0.8921}&
        \footnotesize \centering {0.9125}&
        \footnotesize \centering {0.8945}&
        \footnotesize \centering {0.9585}&
        \footnotesize \centering \textbf{0.9682}&
        \footnotesize \centering {0.9594}&
        \footnotesize {0.9269} \\
         \cline{3-10}
        & &\footnotesize  Mask R-CNN&
        
        \footnotesize \centering \textbf{0.9314}&
        \footnotesize \centering \textbf{0.9504}&
        \footnotesize \centering \textbf{0.9408}&
        \footnotesize \centering {0.9523}&
        \footnotesize \centering {0.9489}&
        \footnotesize \centering {0.9506}&
        \footnotesize 
        \textbf{0.9417} \\
        \cline{2-10}
        \hline
        
    \end{tabularx}
    \caption{Table Structural Segmentation Performance on cross-dataset evaluations. In this table, $\dagger$ represents the only approach that did not utilize the optimized anchors and the results are taken from DeepTabStR \cite{b33} in order to have a direct comparison. Reset of the models operate on optimized anchors. }
    \label{tab:cross_dataset}
\end{table*}


\section{Datasets}
\label{sec:datasets}
We have used two publicly available table structure recognition datasets to conduct the experiments. The particulars of these datasets are explained below.

\subsection{ICDAR-2013}

ICDAR-2013 \cite{b18} dataset has been used to standardize the state-of-the-art results for the task of table detection and table structure recognition \cite{b27, b32}. There is a total of 238 pages in the dataset out of which 156 contains tabular structures. Originally, the dataset contains labels for cells in a table. However, we have used the transformed version of the dataset \footnote{ICDAR-2013 dataset is publicly available at : https://bit.ly/2RLgFYu} published by Siddiqui \textit{et al.} \cite{b32}. The authors have converted the cell-based annotations into the corresponding labeling for rows and columns. We have used the identical test split as employed by Schreiber \textit{et al.} \cite{b27} in order to implement a direct comparison against the similar approaches \cite{b27,b32,b33}. A sample tabular image is illustrated in Figure \ref{fig:table_str_confusion}.

\subsection{TabStructDB}

A Page Object Detection (POD) competition was arranged in ICDAR 2017. The task of this competition was to detect graphical page objects in documents like a table, figures, charts, and equations \cite{b55}. By leveraging this dataset, Siddiqui \textit{et al.} \cite{b32} has published a new dataset for table structure recognition known as TabStructDB \footnote{TabStructDB is publicly available at: https://bit.ly/2XonOEx}. The dataset contains structural information of each table present in the ICDAR-2017 POD dataset. In the dataset, each complete row has been annotated separately regardless of the textual region in order to maintain consistency. Hence, making this dataset ideal for the object detection approach.  To keep the coherence with the ICDAR-2017 POD dataset, the same dataset split has been preserved. The dataset comprised of 731 tabular regions for training whereas 350 tabular regions are preserved for the testing part. A sample tabular image is illustrated in Figure \ref{fig:table_str_confusion}.

\section{Evaluation}
\label{sec:evaluation}


\newcolumntype{t}{>{\hsize=.7\hsize}X}

\begin{table*}
    \centering
    \large
    \setlength\tabcolsep{5pt} 
    \setlength\extrarowheight{5pt}
    \begin{tabularx}{\linewidth}{|t|v|v|k|v|v|k|v|v|k|}
        \specialrule{.2em}{.1em}{.1em}
        \multirow{2}{2cm}{\textbf{Model}}&
        \multicolumn{3}{c|}{\textbf{Row}}&
        \multicolumn{3}{c|}{\textbf{Column}}&
        \multicolumn{3}{c|}{\textbf{Average}}\\
        \cline{2-10}
        & \small \textbf{Precision}&
        \small \textbf{Recall}&
        \small \textbf{F-Measure}&
        \small \textbf{Precision}&
        \small \textbf{Recall}&
        \small \textbf{F-Measure}&
        \small \textbf{Precision}&
        \small \textbf{Recall}&
        \small \textbf{F-Measure}\\
        \hline
        
        \footnotesize {DeepDeSRT}  \cite{b27} &
        \normalsize \centering \textbf{-}&
        \normalsize \centering \textbf{-}&
        \normalsize \centering \textbf{-}&
        \normalsize \centering \textbf{-}&
        \normalsize \centering \textbf{-}&
        \normalsize \centering \textbf{-}&
        \normalsize \centering 0.9593&
        \normalsize \centering 0.8736&
        \normalsize 
        0.91444 \\
        \hline
        \footnotesize {TableNet \cite{b28} } &
        \normalsize \centering \textbf{-}&
        \normalsize \centering \textbf{-}&
        \normalsize \centering \textbf{-}&
        \normalsize \centering \textbf{-}&
        \normalsize \centering \textbf{-}&
        \normalsize \centering \textbf{-}&
        \normalsize \centering 0.9307&
        \normalsize \centering 0.9001&
        \normalsize 
        0.9151 \\
        \hline
        \footnotesize {DeepTabStR \cite{b33}} &
        \normalsize \centering 0.8845&
        \normalsize \centering 0.8945&
        \normalsize \centering 0.8861&
        \normalsize \centering \textbf{0.9688}&
        \normalsize \centering {0.9630}&
        \normalsize \centering \textbf{0.9655}&
        \normalsize \centering 0.9319&
        \normalsize \centering 0.9308&
        \normalsize 
        0.9298 \\
        \hline
        \footnotesize Siddiqui \textit{et al.} \cite{b32} &
        \normalsize \centering 0.9233&
        \normalsize \centering 0.9203&
        \normalsize \centering 0.9190&
        \normalsize \centering 0.9281&
        \normalsize \centering 0.9341&
        \normalsize \centering 0.9288&
        \normalsize \centering 0.9257&
        \normalsize \centering 0.9272&
        \normalsize 
        0.9239 \\
        \hline
        \footnotesize \textbf{Proposed System \newline (With post- \newline processing)} &
        \normalsize \centering \textbf{0.9468}&
        \normalsize \centering \textbf{0.9452}&
        \normalsize \centering \textbf{0.9460}&
        \normalsize \centering 0.9605&
        \normalsize \centering \textbf{0.9659}&
        \normalsize \centering 0.9632&
        \normalsize \centering \textbf{0.9537}&
        \normalsize \centering \textbf{0.9556}&
        \normalsize 
        \textbf{0.9546} \\
        \hline
        \footnotesize \textbf{Proposed System \newline (Without post- \newline processing)} &
        \normalsize \centering {0.9106}&
        \normalsize \centering {0.9326}&
        \normalsize \centering {0.9206}&
        \normalsize \centering 0.9605&
        \normalsize \centering \textbf{0.9659}&
        \normalsize \centering 0.9632&
        \normalsize \centering {0.9355}&
        \normalsize \centering {0.9441}&
        \normalsize 
        {0.9419} \\
         \specialrule{.2em}{.1em}{.1em} 
    \end{tabularx}
    \caption{Table structural recognition performance comparison on ICDAR-2013 dataset. Outstanding results are highlighted. Our proposed system has out-smarted the prior approaches even without the post-processing included.}
    \label{tab:comparison_with_other}
\end{table*}

In order to compare our approach with state-of-the-art methods \cite{b27,b32,b33}, we have used the identical evaluation metrics which are explained below:

\subsection{Intersection over Union (IoU)}
Intersection over Union is a famous evaluation metric used to determine the performance of object detection algorithms. It defines as a measure of a predicted region overlapped with the actual ground truth region. We have used an IoU threshold of 0.5 for the detections. The formula for computing IoU is mentioned below :
\begin{equation}
\frac{\text{Area of Overlap region}} {\text{Area of Union region}}
 \end{equation}

\subsection{Precision}
Precision is defined as the ratio of correctly predicted region and the total predicted region. The formula for precision is explained below :
\begin{equation}
\frac{\text{Predicted area in ground truth}} {\text{Total area of predicted region}}
 = \frac{\text{TP}}{\text{TP $+$ FP}}
\end{equation}
\subsection{Recall}
Recall is calculated as the ratio of correctly predicted region and the total ground truth region. The formula for recall is explained as follows : 
\begin{equation}
\frac{\text{Predicted area in ground truth}} {\text{Total area of ground truth region}}
 = \frac{\text{TP}}{\text{TP $+$ FN}}
 \end{equation}
 
\subsection{F-Measure}
Harmonic mean of precision and recall is known as the F-meausre. The formula for F-measure is :
\begin{equation}
\frac{\text{$2 \times$ Precision $\times$ Recall}} {\text{Precision $+$ Recall}}
 \end{equation}

It is important to understand that the precision, recall, and F-measure are calculated independently for each document followed by taking an average over the complete dataset. This evaluation criterion reduces the bias from a single document containing several rows and columns.

\subsection{Experiments}

As described in Section \ref{sec:datasets}, we have evaluated our proposed approach on the two publicly available datasets i.e. ICDAR-2013 table structure recognition dataset and TabStructDB. Apart from evaluating the datasets on their respective test sets, we have appraised the generalization potential of our approach through the cross-dataset evaluation. The obtained results are highlighted in Table .

\subsubsection{ICDAR-2013}
Since we are using the modified version of the ICDAR-2013 dataset and we report results on the basis of rows and columns, our approach cannot be directly compared with any of the participants of ICDAR-2013 table competition \cite{b18} and other methods operating on cell-level information.
Hence, we compare our approach with the other image-based models who has reported results on rows and columns. To enable the direct comparison with those approaches, we have used the same train/test split proposed by Schreiber \textit{et al.} \cite{b27}.

Table \ref{tab:comparison_with_other} summarizes the results of image-based table structure recognition methods on ICDAR-2013 dataset. Results depict that our proposed approach both (with and without the involved processing method) has outperformed the previous state-of-the-art techniques with an average F-Measure of almost $0.95$. Although results on the column detection of our model are comparable with the DeepTabStR \cite{b33}, our anchor optization method has surpassed the performance of row detections resulting in noticeable improvement on the average results. For cross-dataset evaluation, the model is trained on the TabStructDB dataset and tested on the complete as well as the test set of the ICDAR-2013 dataset. The average F-measure of almost $0.74$ for the test set and almost $0.77$ for the complete dataset in Table \ref{tab:cross_dataset} present the diversity between the two datasets and indicate that there is still room in generalizing the system over various datasets.

\begin{figure*}[ht]
    \includegraphics[width =\linewidth]{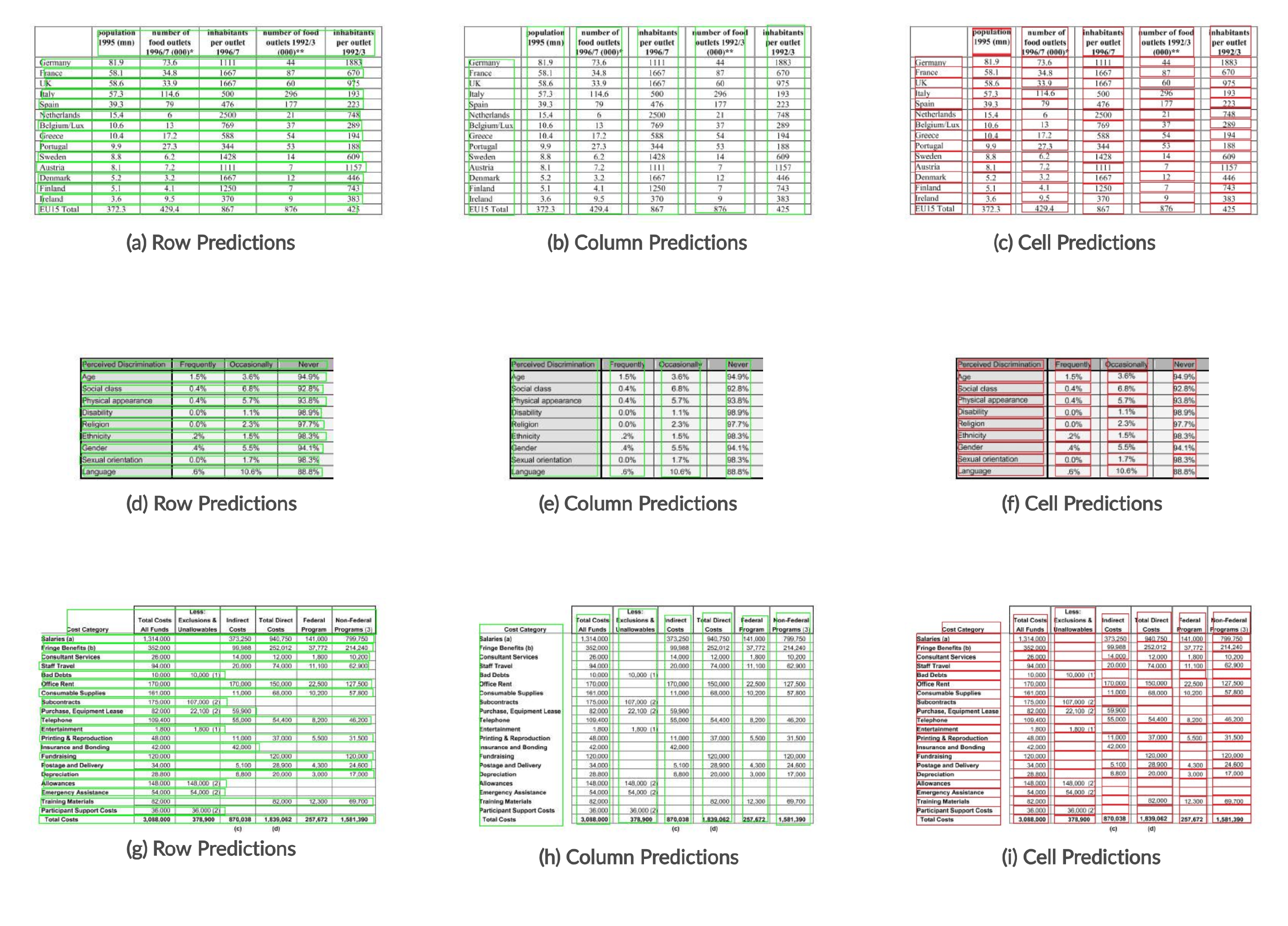}
    \caption{Correctly Recognized Table Structures.}
    \label{fig:coorect_results}
\end{figure*}

\begin{figure*}[ht]
    \includegraphics[width =\linewidth]{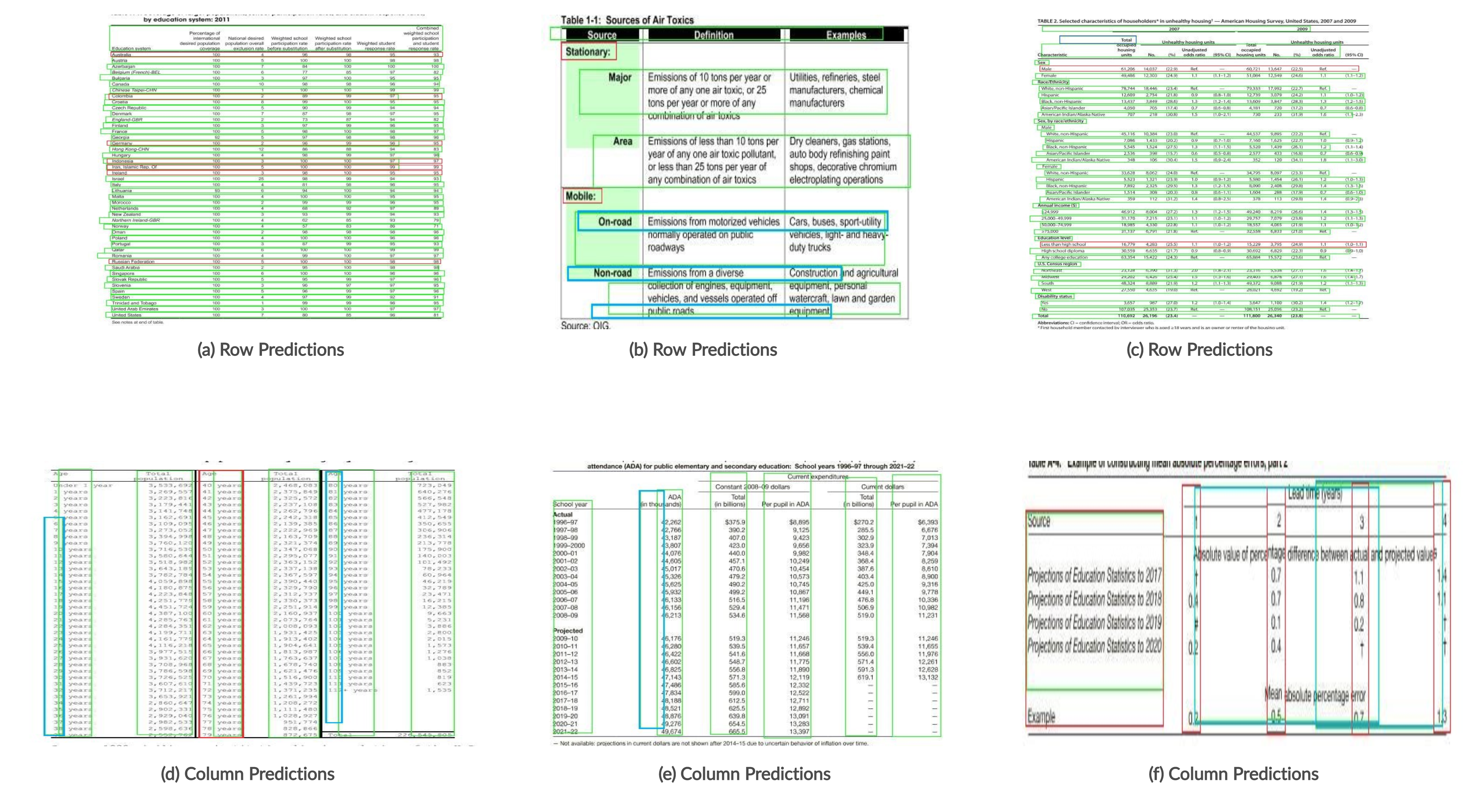}
    \caption{Examples representing incorrectly recognized row and column detection. Green colour shows true positives, blue colour depicts false positives and red colour portrays false negatives for both rows and columns.}
    \label{fig:incorrect_result_rows}
\end{figure*}

Figure \ref{fig:coorect_results} portrays fragments of correctly recognized tabular structures whereas Figure \ref{fig:incorrect_result_rows} depicts some of the cases where rows and columns are not properly detected by the system. In case of incorrect recognition, the model fails to detect few rows in Figure \ref{fig:incorrect_result_rows}(a) and \ref{fig:incorrect_result_rows}(c) because of having several rows with small width in a document image. In another case in Figure \ref{fig:incorrect_result_rows}(b), the system was unable to recognize the row spanning in multiple lines.
Although most of the columns are correctly detected by the model, there are few instances where the system either not capture the whole column area or merge multiple small columns into a single column (Figure \ref{fig:incorrect_result_rows}(d-f)). 

\begin{table*}
    \centering
    
    \normalsize
    \setlength\tabcolsep{5pt} 
    \setlength\extrarowheight{5pt}
    \begin{tabularx}{\linewidth}{|t|v|v|k|v|v|k|k|}
        \specialrule{.2em}{.1em}{.1em} 
        \captionsetup{font=scriptsize}
        \centering \multirow{2}{2cm}{\textbf{Model}}&
        \multicolumn{3}{c|}{\textbf{Row}}&
        \multicolumn{3}{c|}{\textbf{Column}}&
        \multirow{2}{2cm}{\textbf{Average F-Measure}}\\
        \cline{2-7}
        & \small \centering\textbf{Precision}&
        \small \centering\textbf{Recall}&
        \small \centering\textbf{F-Measure}&
        \small \centering\textbf{Precision}&
        \small \centering\textbf{Recall}&
        \small \centering\textbf{F-Measure}&
        \small \textbf{ }\\
        \hline
        \normalsize \centering{DeepTabStR \cite{b33}} &
        \normalsize \centering 0.9093&
        \normalsize \centering 0.9404&
        \normalsize \centering 0.9186&
        \normalsize \centering \textbf{0.9560}&
        \normalsize \centering \textbf{0.9628}&
        \normalsize \centering \textbf{0.9559}&
        \normalsize 
        0.9372 \\
        \hline
        \normalsize \centering\textbf{Proposed System} &
        \normalsize \centering \textbf{0.9314}&
        \normalsize \centering \textbf{0.9504}&
        \normalsize \centering \textbf{0.9408}&
        \normalsize \centering 0.9523&
        \normalsize \centering 0.9489&
        \normalsize \centering 0.9506&
        
        \normalsize 
        \textbf{0.9457} \\
         \specialrule{.2em}{.1em}{.1em} 
    \end{tabularx}
    \caption{Table structural recognition performance comparison on TabstructDB dataset. Outstanding results are highlighted. }
    \label{tab:comparison_reults_tabStruct}
\end{table*} 


\subsubsection{TabStructDB}

Along with the ICDAR-2013 dataset, we have also compared our approach to the TabStructDB dataset. It is evident in the Table \ref{tab:comparison_reults_tabStruct} that our proposed system has outperformed the baseline results established by the DeepTabStR \cite{b33} with an average F-Measure of 0.9417. It is important to mention that since we have trained our models for rows and columns separately, we have compared our results with theirs achieved on separate training methods with the same train/test split of the dataset. 

For the cross-dataset evaluation, a noticeable fall in performance can be perceived in Table \ref{tab:cross_dataset} when the system (trained on ICDAR-2013) is evaluated on TabStructDB complete set and test-set. One of the main reasons for this decline is the disparity in the annotation scheme. The annotations of ICDAR-2013 are limited to textual regions only while TabStructDB has been labeled with complete rows and columns without considering the textual regions at all. Since this is an unrealistic scenario, we have not applied the proposed post-processing method while evaluating the performance of our system on the TabStructDB dataset.

\section{Conclusion and Future Work}
\label{sec:conclusion}
We have proposed a novel approach that employs object detection as a base and adds intelligent automatic estimation of anchor boxes that are suitable for table structure recognition. In this paper, we exhibit that current object detectors that have already shown remarkable improvements in resolving the problem of table detection \cite{b27,b41}, are also extremely effective in improving the performance of table structure recognition systems. We have adopted the anchor optimization technique that facilitates the object detection process with a faster network convergence, a simple post-processing method at the end has further enhanced the performance of our table recognition system. The achieved results have evidently outperformed the state-of-the-art image-based table structure recognition system on the publicly available ICDAR-2013 dataset with an average F-Measure of $95.05\%$ and surpassed the baseline results on the publicly available TabStructDB dataset with an average F-Measure of $94.17\%$. The obtained results recommend the idea of exploiting object detectors for table recognition systems.

Although our proposed method is nearly applicable to all of the document images, some exceptional cases do exist. Hence, better post-processing methods should be developed. Our model had a hard time detecting rows spanning for multiple lines in the table. An interesting direction could be to detect the cells directly instead of rows and columns. Instead of using the traditional convolutional neural networks, recently proposed  CoordConv \cite{b57} could also be exploited in the object detection algorithms in order to provide the system with extra contextual information. This paper tackles the problem of table structure recognition in business-like scanned document images. It would be interesting to examine this approach for the datasets that contains historical document images such as ICDAR-2019 (cTDaR) \cite{b32}.



\clearpage

\vspace{-1,5 cm}
\begin{IEEEbiography}[{\includegraphics[width=1in,height=1.25in,clip,keepaspectratio]{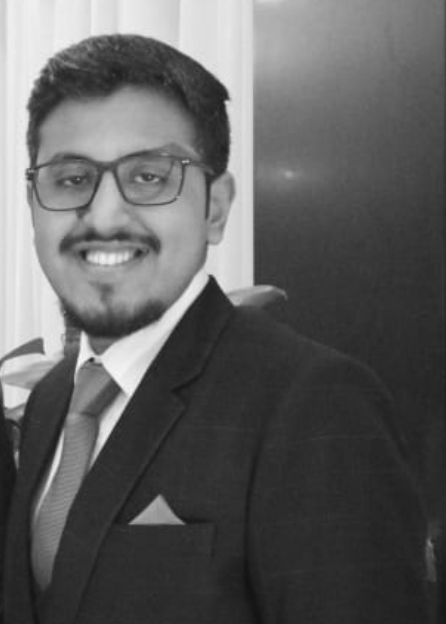}}]{Khurram Azeem Hashmi} received his bachelor’s degree in computer science from the National University of Computer and Emerging Sciences, Pakistan in 2016, and the M.S. degree from the Technical University of Kaiserslautern. He is currently pursuing a Ph.D. degree with the German Research Center for Artificial Intelligence (DFKI GmbH) and the Technical University of Kaiserslautern, under the supervision of Prof. Dr. Didier Stricker. His areas of interest include deep learning for computer vision specifically in object detection and activity recognition. He is also interested in the area of pattern recognition and document analysis. Previously, he has worked in the field of document layout understanding and post-OCR error corrections. He is also a Reviewer for IEEE Access.
\end{IEEEbiography}

\vskip 0pt plus -1fil

\begin{IEEEbiography}[{\includegraphics[width=1in,height=1.25in,clip,keepaspectratio]{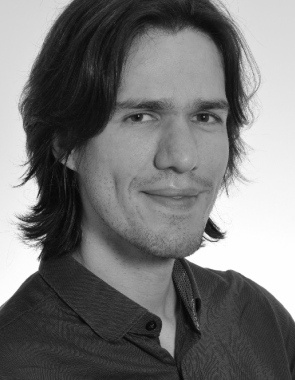}}]{Marcus Liwicki} received his M.S. degree in Computer Science from the Free University of Berlin, Germany, in 2004, his PhD degree from the University of Bern, Switzerland, in 2007, and his habilitation degree at the Technical University of Kaiserslautern, Germany, in 2011. Currently he is chaired professor at Luleå University of Technology and a senior assistant in the University of Fribourg. His research interests include machine learning, pattern recognition, artificial intelligence, human computer interaction, digital humanities, knowledge management, ubiquitous intuitive input devices, document analysis, and graph matching. He is a member of the IAPR, editor or regular reviewer for international journals, including IEEE Transactions on Pattern Analysis and Machine Intelligence, IEEE Transactions on Audio, Speech and Language Processing, International Journal of Document Analysis and Recognition (editor), Frontiers of Computer science (editor), Frontiers in Digital Humanities (editor), Pattern Recognition, and Pattern Recognition Letters. He is a member of governing board the International Graphonomics Society and a member of the International Association for Pattern Recognition where he is Vice president of the Technical Committee 6. He chaired several International Workshops on Automated Forensic Handwriting Analysis and the International Workshop on Document Analysis Systems 2014. Furthermore he serves as program committee member and reviewer of various International Conferences and workshops in the area of Computer Vision, Pattern Recognition and Document Analysis as well as Machine Learning and E-Learning.
\end{IEEEbiography}
\vskip 0pt plus -1fil
\begin{IEEEbiography}[{\includegraphics[width=1in,height=1.25in,clip,keepaspectratio]{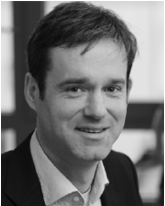}}]{Didier Stricker} is professor with the University of Kaiserslautern and scientific director with the “German Research Center for Artificial Intelligence” (DFKI) in Kaiserslautern where he leads the research department Augmented Vision. From 2002 to June 2008, he lead the department “Virtual and Augmented Reality” at the Fraunhofer Institute for Computer Graphics (Fraunhofer IGD) in Darmstadt, Germany. In this function, he initiated and participated to many national and international projects in the areas of computer vision and virtual and augmented reality. In 2006, he received the Innovation Prize of the German Society of Computer Science. He serves as reviewer for different European or national research organizations, and is a regular reviewer for the most important journals and conferences in the areas of VR/AR and computer vision.
\end{IEEEbiography}

\vskip 0pt plus -1fil
\begin{IEEEbiography}[{\includegraphics[width=1in,height=1.25in,clip,keepaspectratio]{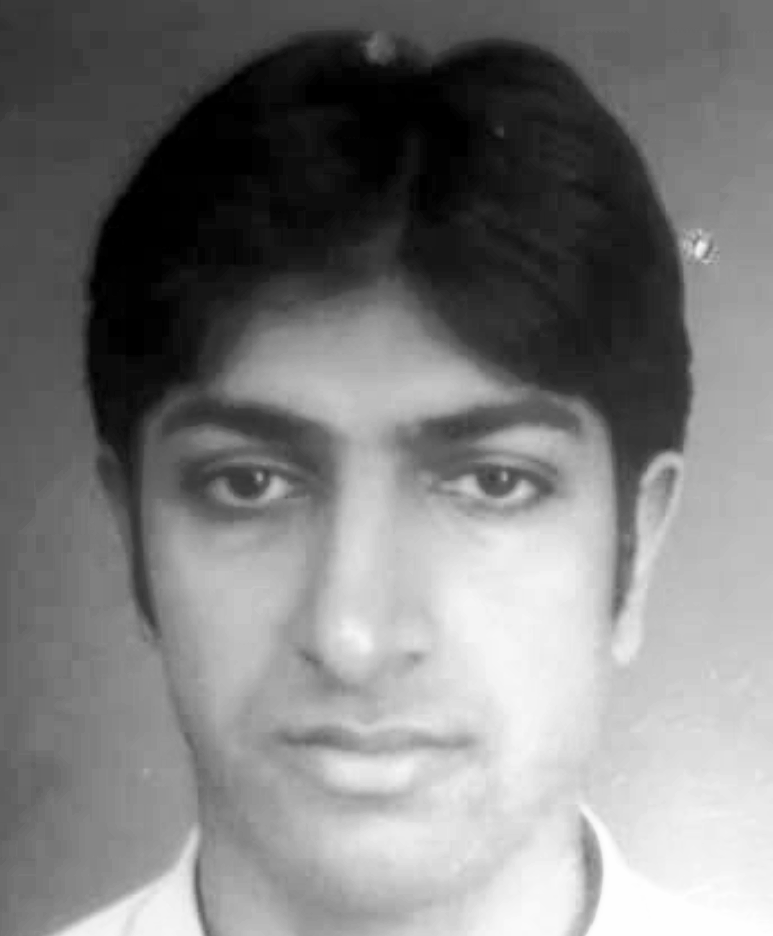}}]{Muhammad Noman Afzal} completed his bachelor's degree in Computer Science from Islamia University of Bahawalpur, Pakistan. He is currently involved in research and development in the area of Artificial Intelligence. He is a  deep learning enthusiast. He has over 7 years of work experience with different types of deep learning techniques. However, he is mostly interested in object detection. His general interests are deep learning in challenging environments. He has also worked with deploying artificial intelligence at the edge. He has vast experience in mobile development. Furthermore, he is involved in academia where he delivers lectures on machine learning.
\end{IEEEbiography}

\vskip 0pt plus -1fil
\begin{IEEEbiography}[{\includegraphics[width=1in,height=1.25in,clip,keepaspectratio]{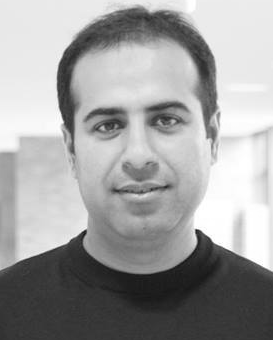}}]{Muhammad Zeshan Afzal} received his Masters degree from the University of Saarland, Germany majoring in Visual Computing in 2010 and his Ph.D. degree from the University of Technology, Kaiserslautern, Germany majoring in Artificial Intelligence in 2016. His research interests include deep learning for vision and language understanding using deep learning.  At an application level, his experience includes generic segmentation framework for natural, human activity recognition, document and, medical image analysis, scene text detection, and recognition, on-line and off-line gesture recognition. Moreover, a special interest in recurrent neural networks for sequence processing applied to images and videos. He also worked with numerics for tensor valued images. He worked both in the industry (Deep Learning and AI Lead Insiders Technologies GmbH) and academia (TU Kaiserslautern). He received the gold medal for the best graduating student in Computer Science from IUB Pakistan in 2002 and secured a DAAD(Germany) fellowship in 2007. He is a member of IAPR. 
\end{IEEEbiography}

\EOD

\end{document}